\documentclass{article}

\usepackage{arxiv}

\usepackage[utf8]{inputenc} 
\usepackage[T1]{fontenc}    
\usepackage{hyperref}       
\usepackage{url}            
\usepackage{booktabs}       
\usepackage{amsfonts}       
\usepackage{nicefrac}       
\usepackage{microtype}      
\usepackage{lipsum}		
\usepackage{graphicx}

\usepackage{doi}
\usepackage{url,hyperref,lineno,microtype,subcaption}
\usepackage[onehalfspacing]{setspace}
\usepackage{bm}
\usepackage{amsmath}
\usepackage{multirow}

\usepackage{amssymb}
\usepackage{bm}
\usepackage{gensymb}
\usepackage{amsmath}
\usepackage{algorithm}
\usepackage{algpseudocode}
\usepackage{booktabs}
\usepackage{multirow}
\usepackage{enumitem} 
\newlist{condenum}{enumerate}{1} 
\setlist[condenum]{label=\bfseries Condition \arabic*., 
                   ref=\arabic*, wide}
\usepackage{changepage}

\title{Multiple-object Grasping Using a Multiple-suction-cup Vacuum Gripper in Cluttered Scenes}

\author{ {}\\
{Ping Jiang$^{*}$, Junji Oaki, Yoshiyuki Ishihara, and Junichiro Ooga}\\
Corporate Research \& Development Center\\
Toshiba Corporation\\
1, Komukai-Toshiba-cho, Saiwai-ku, Kawasaki 212-8582, Japan.\\
	\texttt{ping2.jiang@toshiba.co.jp}\\
}

\renewcommand{\shorttitle}{\textit{arXiv} Template}

\hypersetup{
pdftitle={A template for the arxiv style},
pdfsubject={q-bio.NC, q-bio.QM},
pdfauthor={Ping Jiang},
pdfkeywords={bin picking, multiple-object grasping, grasp planner,  vacuum gripper},
}

\begin{document}
\maketitle

\begin{abstract}
Multiple-suction-cup grasping can improve the efficiency of bin picking in cluttered scenes. In this paper, we propose a grasp planner for a vacuum gripper to use multiple suction cups to simultaneously grasp multiple objects or an object with a large surface. To take on the challenge of determining where to grasp and which cups to activate when grasping, we used 3D convolution to convolve the affordable areas inferred by neural network with the gripper kernel in order to find graspable positions of sampled gripper orientations. The kernel used for 3D convolution in this work was encoded including cup ID information, which helps to directly determine which cups to activate by decoding the convolution results. Furthermore, a sorting algorithm is proposed to find the optimal grasp among the candidates. Our planner exhibited good generality and successfully found multiple-cup grasps in previous affordance map datasets. Our planner also exhibited improved picking efficiency using multiple suction cups in physical robot picking experiments. Compared with single-object (single-cup) grasping, multiple-cup grasping contributed to $1.45\times$, $1.65\times$, and $1.16\times$ increases in efficiency for picking boxes, fruits, and daily necessities, respectively.
\end{abstract}

\keywords{bin picking \and grasp planning \and suction grasp \and graspability \and deep learning}

\section{Introduction}
\begin{figure}[hbtp]
 \centering
 \includegraphics[keepaspectratio, scale=0.3]{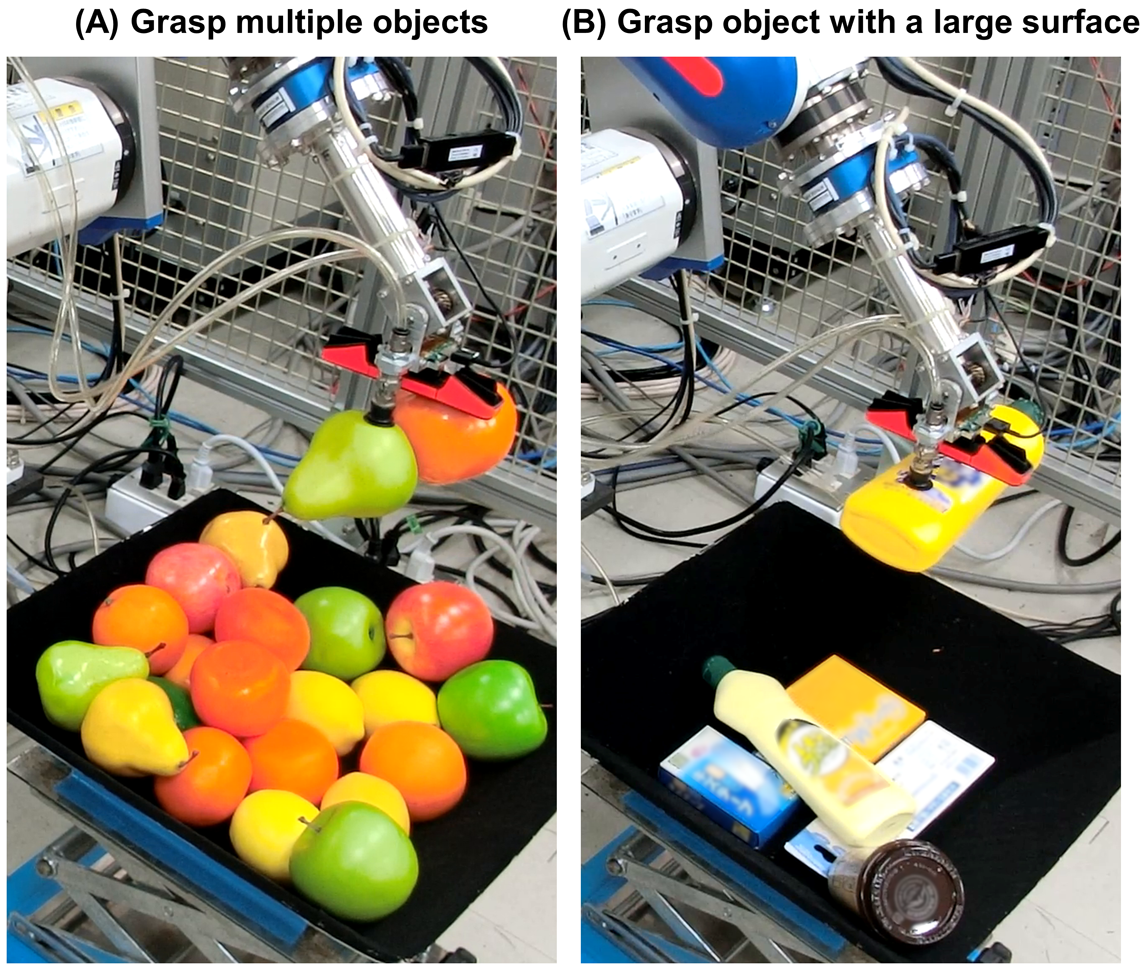}
 \caption{Using multiple suction cups to grasp (A) multiple objects or (B) an object with a large surface.}
 \label{fig1}
\end{figure}

With the growth of e-commerce, demand for automation of bin picking by robots in warehouses has become high \cite{1}, particularly in Japan since the country is faced with a labor shortage due to its aging society. Covid-19 has made the situation worse since the picking task in warehouses is not amenable to telework. Most state-of-the-art robotic picking systems have focused on single-object grasping. To further improve the efficiency of these systems, simultaneous grasping of multiple objects might reduce the number of pick attempts to improve the picking speed as shown in Fig. \ref{fig1} (A). In addition, a robot can more stably grasp and hold objects that have a large surface by using multiple suction cups to grasp the object as in Fig. \ref{fig1} (B). 

Multiple-object grasp planning for jaw or multi-finger grippers has previously been proposed under various conditions, such as in well-organized scenes \cite{22,23}, rearranged scenes \cite{27}, and cluttered scenes \cite{25,26,28,29}. These studies demonstrated that multiple-object grasping could improve picking speed. However, few studies have examined multiple-object grasping by a vacuum gripper with multiple suction cups. Most studies infer the single-object grasp point for a gripper with only a single suction cup using direct or indirect methods. Direct methods \cite{5,6,7} use deep convolutional neural networks to directly infer the grasp point, while indirect methods \cite{9,38,39} first infer the affordance map, which is a pixel-wise map indicating the graspability score for a single-cup vacuum gripper at each pixel, and then find the optimal grasp point in the map. Given that the affordance map contains all possible grasp points for a single suction cup, if all cups in a vacuum gripper have the same geometry (e.g., cup radius) and dynamics (e.g., suction force limit and friction coefficient), then we can search for a gripper pose where the center positions of at least two of the cups are located at non-zero pixels in the affordance map and satisfy the conditions described in Section 4 for grasping multiple objects or an object with a large surface.

In this study, we propose an affordance-map-based grasp planner for a multiple-suction-cup vacuum gripper to grasp multiple objects or grasp an object with a large surface. We propose a 3D-convolution-based method, which takes advantage of the suction affordance map inferred by our prior work, suction graspability U-Net++ (SG-U-Net++) \cite{39}, to search for a gripper pose capable of grasping multiple objects or an object with a large surface. Furthermore, unlike the control of a jaw gripper in which all fingers of the gripper are usually controlled to open or close simultaneously, the suction cups need to be controlled separately. Therefore, we designed a kernel that included encoded cup ID to determine which suction cup to activate. Furthermore, as there might be many candidates for multiple-suction-cup grasping, we propose an evaluation metric for finding the optimal grasp among candidates. The proposed grasp planner was validated on previous affordance datasets and by physical robot experiments.

In short, the contributions of this work include the following:
\begin{itemize}
\item A 3D-convolution-based grasp planner for a vacuum gripper with multiple cups to grasp multiple objects or an object with a large surface.
\item Control of suction cup activation by incorporating a convolution kernel including the encoded cup ID.
\item A robotic picking system with hybrid planner that performs multiple-suction-cup grasp planning preferentially, and switches to single-object grasp planning when there are no solutions.
\item A sorting algorithm for finding the optimal grasp for multiple-cup grasping.
\item Validation of the grasp planner on previous affordance datasets including Suction FCN \cite{9}, SuctionNet-1Billion\cite{38}, and SG-U-Net++\cite{39}.
\item Experiments on picking boxes, fruits, and daily necessities by a vacuum gripper with two cups and a comparison of multiple- and single-cup grasping results.
\end{itemize}

\section{Related works}
\subsection{Single-object grasping based on an affordance map}
A pixel-wise affordance map contains grasp quality at each pixel when the robot grasps the object in the corresponding pose. Unlike end-to-end deep learning, which has been used to directly predict grasp configurations such as a rotated bounding box \cite{2,3,4} for a jaw gripper or a suction point for a vacuum gripper with only a single cup \cite{5,6,7}, affordance learning has the advantages that the neural network model can be anchor-free and there is no need to sample candidate grasps as in \cite{8}. Zeng et al. \cite{9} were one of the first researchers to apply pixel-wise affordance learning to bin picking for four multiple-motion primitives of a hybrid robotic hand with both a jaw and single suction cup. They used a manually annotated affordance dataset to train fully convolutional networks (FCNs). The precision and generalizability of FCNs were further improved by \cite{10, 11}. Another representative work is by Morrison et al. \cite{12}, who generated affordance and pose maps from the rotated bounding box and designed a Generative Grasp CNN (GG-CNN) to directly infer pixel-wise grasp pose and quality. To learn the grasp of a jaw gripper, many researchers \cite{13,14,15,16,17} later used similar methods to generate affordance map datasets from grasp configuration annotations represented by a rotated rectangle (e.g., the Cornell Grasp Dataset \cite{2} and Jacquard Dataset \cite{18}).

However, these studies required real images and an expert to perform pixel-wise grasp affordance annotation. To reduce dataset generation costs, datasets are generated in a physical simulator where affordance is evaluated using a designed contact model (e.g., the quasi-static spring model used in Dex-net \cite{19}) on a previously synthesized image. Recently, a similar contact model was used by Cao et al. \cite{38} to generate a larger suction pixel-wise affordance (seal score) dataset. However, these studies required that contact model parameters were determined properly and treated a vacuum gripper with only a single cup.

In the present study, we used our previously proposed SG-U-Net++ \cite{39} to infer the pixel-wise suction affordance map. SG-U-Net++ was trained on a synthesized dataset annotated by an analysis-based method and was competitive with method trained on a dataset annotated by a contact model. We propose a grasp planner for multiple-suction-cup grasping that takes advantage of the predicted affordance map.

\subsection{Multiple-object grasping}
Most studies have focused on multiple-object grasping using a multi-finger gripper. Grasp conditions have been analyzed for a multi-finger gripper to stably grasp multiple cylinders \cite{20,21}, polyhedral objects \cite{22}, planar objects \cite{23}, and shaped spatial objects \cite{24}. Recent studies have started to use data-driven methods to deal with the multiple-object grasping problem. Shenoy et al. \cite{25} used a deep neural network to infer the number of objects for a three-finger gripper to grasp when digging into a pile of objects. They later proposed a Markov decision-based method to optimize the pick-transfer routines when grasping multiple objects \cite{26}. Sakamoto et al. \cite{27} used mask-RCNN to detect objects and then searched for a gripper pose to push two boxes together in order to simultaneously grasp them. A similar push-grasp task was studied by Agboh et al. for grasping multiple arbitrary convex polygonal objects under frictional and frictionless contact conditions between the objects \cite{28,29}. They proposed MOG-Net for inferring the max number of objects that a two-finger gripper could grasp by a sampled pose. However, the set of objects in these studies was still simple, and simultaneous grasping of objects with more complicated shapes (e.g., daily necessities) is required for more general applications (e.g., picking in warehouses). Mucchiani et al. \cite{30} designed a novel end-effector to sequentially grasp multiple objects with complicated shapes. Yao et al. \cite{31} proposed a human-like grasp synthesis algorithm to achieve sequential multiple-object grasping. However, these studies grasped multiple objects sequentially rather than simultaneously.

For multiple-suction-cup grasping, most studies treat grasping a single object rather than multiple objects using multiple cups. Mantriota \cite{32} analyzed the suction force and friction coefficient to grasp and hold a large object by a four-cup vacuum gripper. Koz{\'a}k \cite{34} et al. used a deep neural network to estimate the pose of a round part and then used a six-cup vacuum gripper to grasp it. Tanaka et al. \cite{35} designed a two-surface vacuum gripper in which each surface was equipped with multiple cups. They used a gripper to simultaneously suck two surfaces of a large box to improve the stability of grasping and holding. Leitner et al. \cite{36} used a gripper with two different shaped cups to grasp an object on a shelf. These studies used a multiple-cup vacuum gripper to grasp a single object more stably. Kessens \cite{33} et al. mounted a four-cup vacuum gripper on a drone to achieve sequential multiple-object grasping in the air, but found that simultaneous grasping was challenging. Islam et al. \cite{37} proposed a planner for an unloading task in which the robot used a multiple-suction-cup vacuum gripper to simultaneously grasp and unload multiple cardboard boxes, but it was difficult to apply the planner to objects with complicated shapes in a cluttered scene. To our knowledge, the present study is the first to propose a grasp planner for simultaneously grasping multiple objects using a multiple-suction-cup vacuum gripper. The planner can also find gripper poses for stably grasping  large objects with multiple cups.

\section{Problem statement}
This study focuses on the bin picking task in cluttered scenes. The robot is required to pick multiple objects or an object with a large surface using multiple suction cups, and then to place them in a tote. 

\subsection{Assumption}
\begin{figure}[hbtp]
 \centering
 \includegraphics[keepaspectratio, scale=0.35]{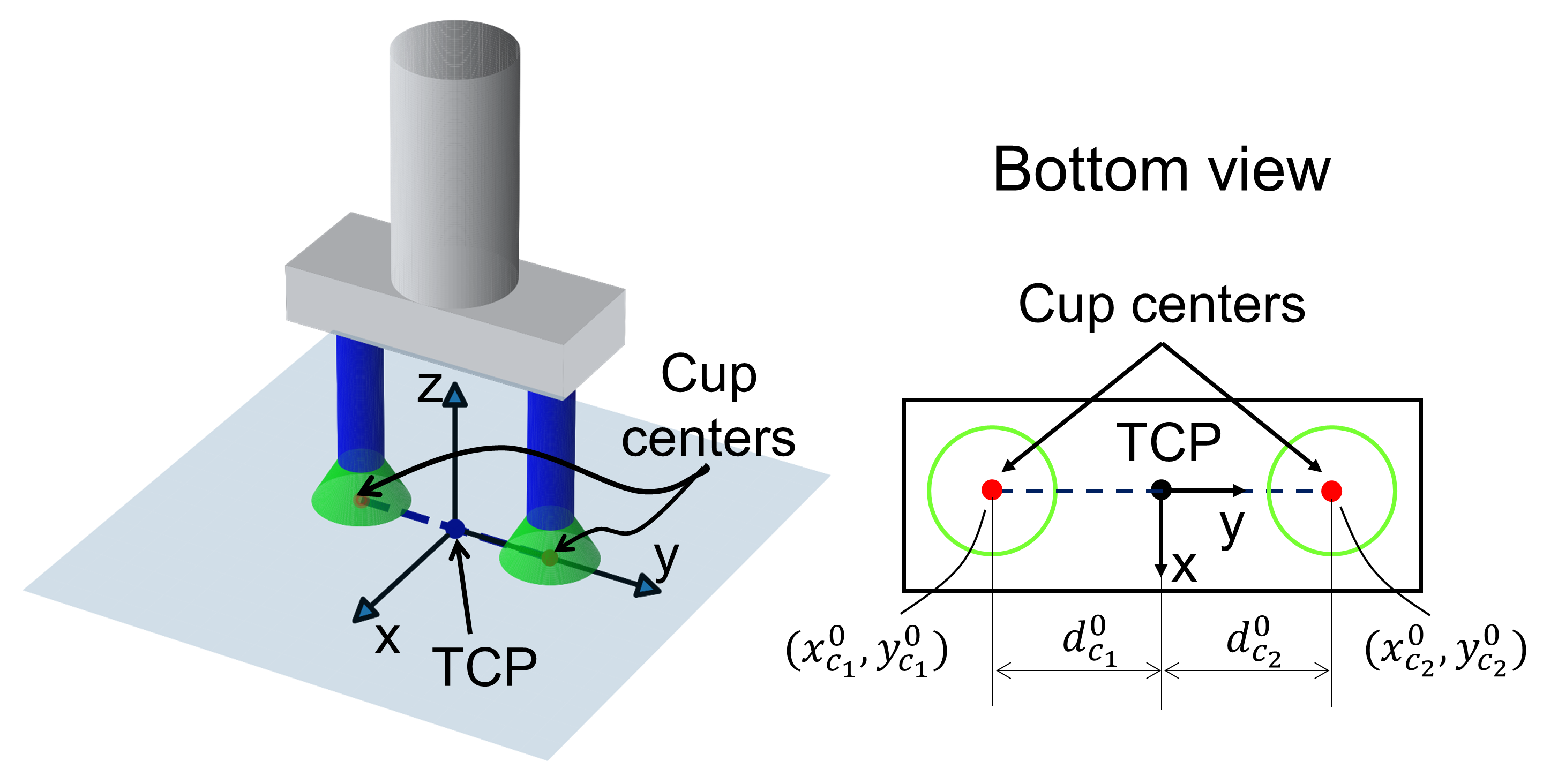}
 \caption{Example of a vacuum gripper with two cups. TCP is the gripper tool center point. $d_{c_i}^0$ is the distance from the $i$th cup to TCP. ($x^0_{c_i}, y^0_{c_i}$) is the center position of the $i$th cup in gripper local coordinates. }
 \label{fig2}
\end{figure}

We assume a suction vacuum gripper with multiple suction cups where all cups have the same specifications (e.g., the right side of Fig. \ref{fig2}, in which both cups have the same shape, size, and suction force limits). In addition, the gripper tool center point (TCP) and all cups are in the same plane (e.g., the left side of Fig. \ref{fig2}, in which the cup center points and TCP are in the same blue plane). 

\subsection{Vacuum gripper state}
The vacuum gripper state $\bm{G}$ consists of gripper position $\bm{P}$, orientation $\bm{O}$, suction cup center positions $\bm{C}$, and cup activation mode $\bm{A}$ as in Eq. (\ref{eq1}). $\bm{P}$ is the position ($x_g, y_g, z_g$) of TCP in world coordinates. $\bm{O}$ is the orientation represented by a ZYZ rotation matrix $(R_z(\theta_{g})R'_y(\phi_{g})R''_z(\gamma_{g})$), where $\theta_{g}$ and $\phi_{g}$ are the azimuthal angle and polar angle of the unit vector of gripper axis z respectively, and $\gamma_{g}$ is the rotation angle around gripper axis z. Note that $sin$ and $cos$ are abbreviated as $c$ and $s$ in the matrix. $\bm{C}$ consists of the center position ($x_{c_i}, y_{c_i}, z_{c_i}$) of each suction cup in world coordinates. $\bm{C}^0$ is the center position ($x_{c_i}^0, y_{c_i}^0, z_{c_i}^0$) of each suction cup in gripper local coordinates (see the right side of Fig. 2). $\bm{A}$ is a one-hot vector representing the activation status ($a_{c_i}$) of each suction cup, where $a_{c_i}$ is 1 if the $i$th cup is activated, and 0 if it is disabled.
\begin{equation} 
\begin{split}
   \bm{G}&=[\bm{P}, \bm{O}, \bm{C}, \bm{A}]\\
   \bm{P}&=[x_g, y_g, z_g]   \\
   \bm{O}&= R_z(\theta_{g})R'_y(\phi_{g})R''_z(\gamma_{g})   \\
    &=
        \begin{bmatrix}
c_{\phi_g} c_{\theta_g} c_{\gamma_g}  -s_{\theta_g} c_{\gamma_g}  & 
-c_{\phi_g} c_{\theta_g} s_{\gamma_g} -s_{\theta_g} c_{\gamma_g} &
s_{\phi_g} c_{\theta_g}\\
c_{\phi_g} s_{\theta_g} c_{\gamma_g} + c_{\theta_g} s_{\gamma_g} & 
-c_{\phi_g} s_{\theta_g} s_{\gamma_g} +c_{\theta_g} c_{\gamma_g} &
s_{\phi_g} s_{\theta_g}\\
-s_{\phi_g} c_{\gamma_g} & s_{\phi_g} s_{\gamma_g} & c_{\phi_g}\\
\end{bmatrix}\\
   \bm{C}&=[[x_{c_1}, y_{c_1}, z_{c_1}], [x_{c_2}, y_{c_2}, z_{c_2}], ..., [x_{c_i}, y_{c_i}, z_{c_i}]]  \\ 
   \bm{A}&=[a_{c_1}, a_{c_2}, ..., a_{c_i}]
\end{split}
\label{eq1}
\end{equation}

\section{Conditions for grasping using multiple suction cups}
\begin{figure}[hbtp]
 \centering
 \includegraphics[keepaspectratio, scale=0.45]{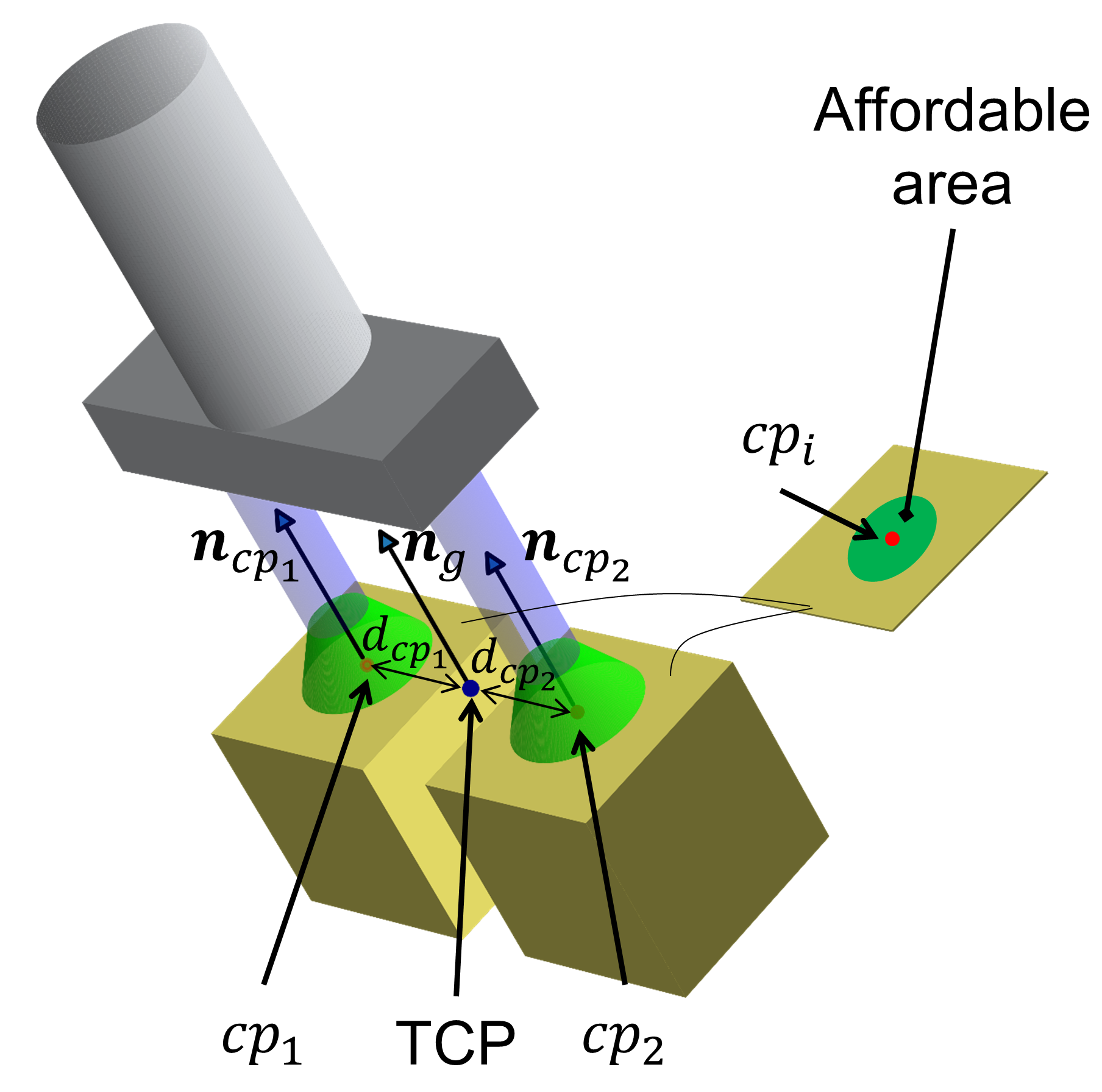}
 \caption{An example of conditions for a vacuum gripper with two cups to grasp two objects. $cp$ is the contact point where the suction cup sucks the surface. Contact points $cp_1$ and $cp_2$ need to be located in the affordable areas of objects. $cp_1$, $cp_2$, and TCP need to be in the same plane perpendicular to $\bm{n}_g$. $\bm{n}_{{cp}_{1}}$ and $\bm{n}_{{cp}_{2}}$ are the normals of the contact points for the left and right cups, respectively. Both  $\bm{n}_{{cp}_{1}}$ and $\bm{n}_{{cp}_{2}}$ need to be parallel to $\bm{n}_g$. $d_{{cp}_{1}}$ and $d_{{cp}_{2}}$ are the distances from the contact points to TCP for the left and right cups in world coordinates, respectively. Both $d_{{cp}_{1}}$ and $d_{{cp}_{2}}$ need to be equal to the distances from the left cup center ($d_{c_1}^0$) and right cup center ($d_{c_2}^0$) to TCP in gripper coordinates.}
 \label{fig3}
\end{figure}

Since all suction cups installed in the gripper are assumed to be the same, the affordance map of each suction cup is the same. Hence, we can find a gripper pose capable of grasping multiple objects or an object with a large surface by multiple cups if the following conditions are satisfied (an example is shown in Fig. \ref{fig3}).

\begin{condenum}
\item At least two of the contact points are located in affordable areas of objects. If the contact points are located in affordable areas of different objects, the gripper can grasp multiple objects. If the contact points are located in the same affordable area, the gripper can grasp a large surface by using multiple cups. \label{cond1}
\item Gripper TCP and all contact points located in affordable areas are in the same plane, which is perpendicular to the unit vector of gripper axis-z ($\bm{n}_g$). \label{cond2}
\item Normals of all contact points located in affordable areas ($\bm{n}_{{cp}_{i}}$) are in the same direction as the unit vector of gripper axis-z ($\bm{n}_g$) as in Eq. (\ref{eq3}). Note that $aff_{cp_i}>0$ indicates that the $i$th contact point is located in an affordable area where its affordance score is non-zero. \label{cond3}
\begin{equation}
  \textrm{arccos}(\bm{n}_{{cp}_{i}} \boldsymbol{\cdot} \bm{n}_g)<\varepsilon_1 \quad \textrm{where}  \quad aff_{cp_i}>0
\label{eq3}
\end{equation}
\item The distance from each contact point located in the affordable areas to TCP in world coordinates ($d_{cp_i}$) needs to be equal to the distance from the corresponding cup center to TCP in local coordinates ($d_{c_i}^0$ in Fig. \ref{fig2}). \label{cond4}
\begin{equation}
  |d_{cp_i}-d_{c_i}^0| < \varepsilon_2 \quad \textrm{where}  \quad aff_{cp_i}>0
\label{eq4}
\end{equation}
\end{condenum}

\section{Multiple-suction-cup grasp planner}
\subsection{Overview of architecture}
\begin{figure*}[hbtp]
 \centering
 \includegraphics[keepaspectratio, scale=0.2]{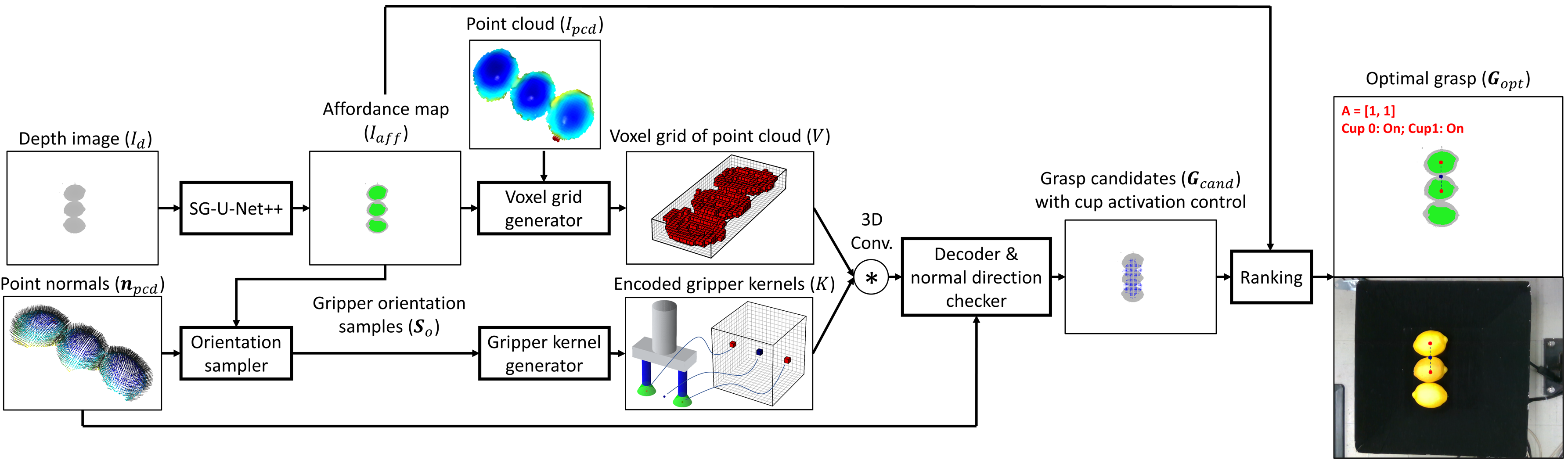}
 \caption{Overall architecture of the multiple-suction-cup grasp planner.}
 \label{fig4}
\end{figure*}

Figure \ref{fig4} and Algorithm \ref{alg1} shows the overall architecture and work flow of our multiple-suction-cup grasp planner. Given a depth image $I_{d}$, our previous work SG-U-Net++ is used to infer the affordance map $I_{aff}$ for a single cup. The voxel grid generator then extracts the point cloud ($I_{pcd}$) affiliated with the affordable areas in the map and downsamples them to a voxel grid ($V$). The orientation generator uses the point normals $\bm{n}_{pcd}$ of extracted points to efficiently generate the gripper orientation samples ($\bm{S}_{O}$). The gripper kernel generator generates 3D encoded gripper kernels ($\mathcal{K}$) including cup ID information. The decoder decodes the result ($ConvRes$) of 3D convolution (3D Conv.) of $V$ over $\mathcal{K}$ and generates the gripper pose candidates ($\bm{G}_{cand}$). The normal direction checker removes candidates where the $\bm{n}_{g}$ and contact point normals are not in the same direction. If $\bm{G}_{cand}$ is successfully found, $\bm{G}_{cand}$ is evaluated and ranked to obtain the optimal grasp ($\bm{G}_{opt}$). Otherwise, if no $\bm{G}_{cand}$ is found, the planner is switched to our previous single-object grasp planner where the position with the highest affordance score is set as the goal and the cup that can reach the goal by the shortest trajectory is selected to grasp the object.

\begin{algorithm}[tb]
\caption{Multiple-suction-cup grasp planner}
\label{alg1}
\hspace*{\algorithmicindent} 
$\begin{aligned}
\textbf{Input: }  &I_{aff}: \textnormal{affordance map} \\
&I_{d}: \textnormal{depth image}\\
&I_{pcd}: \textnormal{point cloud} \\
& l: \textnormal{voxel size} \\
& \bm{C}^0: \textnormal{local cup center positions (see the right side of Fig. 2)}
\end{aligned}$

\hspace*{\algorithmicindent} \textbf{Output: } $G_{opt}$: optimal grasp
\begin{algorithmic}[1]
\State $V \gets \mathrm{GenerateVoxelGrid}(I_{pcl}, I_{aff}, l)$
\State $\bm{n}_{pcd} \gets \mathrm{EstimateNormals}(I_{pcd})$
\State $\bm{S}_{O} \gets \mathrm{SampleGripperOrientation}(\bm{n}_{pcd}, I_{aff})$
\State $\mathcal{K} \gets \mathrm{GenerateEncodedKernels}(\bm{S}_{O}, \bm{C}^0, l)$
\State $ConvRes \gets \mathrm{Conv3D}(V, \mathcal{K})$
\State $\bm{G}_{cand} \gets \mathrm{Decode}(ConvRes)$
\State $\bm{G}_{cand} \gets \mathrm{NormalDirectionCheck}(ConvRes)$
\If {$\mathrm{len}( \bm{G}_{cand})>0$}
\State $\bm{G}_{opt} \gets \mathrm{Ranking}(I_{aff}, \bm{G}_{cand})$
\Else
\State $\textrm{\# Single-object grasp planning}$
\State $\bm{G}_{opt} \gets \mathrm{argmax}(I_{aff})$
\EndIf
\State \Return $\bm{G}_{opt}$
\end{algorithmic}
\end{algorithm}

\subsection{Affordance map inference}

We used SG-U-Net++ from our prior work to generate the affordance map. SG-U-Net++ has a nested U-Net structure, and infers pixel-wise grasp quality and approachability based on a depth image. Refer to \cite{39} for further details. Pixels with non-zero grasp quality score were filtered out to generate an affordance map (green area in affordance map in Fig. \ref{fig4}). 

\subsection{Voxel grid generation}
We used voxel downsampling to generate the binary voxel grid ($V$) of the point cloud. Points located in the affordable areas were extracted and downsampled to a voxel grid with a defined grid size $l$. The voxel grid was further binarized such that if a grid in the voxel grid contained more than 10 points, the grid value would be 1 and else 0. The voxel grid shape is $N_x \times N_y \times N_z$ where $N_x  = \lfloor\frac{B^{max}_x-B^{min}_x}{l}\rfloor$, $N_y = \lfloor\frac{B^{max}_y-B^{min}_y}{l}\rfloor$, $N_z = \lfloor\frac{B^{max}_z-B^{min}_z}{l}\rfloor$. $B^{max}$ and $B^{min}$ are the max and min bounds of the point cloud.

\subsection{Grasp orientation candidate generation}
To satisfy Condition 3 in Section 4, Eq. (\ref{eq3}) needs to be computed for each point normal to sample the gripper orientations. If the size of the input point cloud is large, online sampling will result in high costs in terms of memory usage and computation time. We propose an efficient sampling method for a vacuum gripper by generating an offline normal to gripper orientation map. Since the Cartesian coordinates of a given vector $\bm{vec}$ ($[vec_x, vec_y, vec_z]$) can be represented by the azimuthal  angle $\theta$ and polar angle $\phi$ as in Eq. (\ref{eq5}), all possible normals of contact points can be sampled by an angle interval $\Delta\alpha$ as in Eq. (\ref{eq6}). Meanwhile, as in Eq. (\ref{eq1}), $\bm{n}_g$ (the last column of $\bm{O}$) depends on only $\phi_g$ and $\theta_g$ and has the same representation as Eq. (\ref{eq5}), so $\bm{n}_g$ can be sampled by the same angle interval as in Eq. (\ref{eq7}).
\begin{equation}
\begin{split}
  &\bm{vec}=[vec_x, vec_y, vec_z]=[c_{\theta}s_{\phi}, s_{\theta}s_{\phi},c_{\phi}] \\
  &\theta =  \textrm{arctan}(vec_y, vec_x) \\
  &\phi = \textrm{arccos}(vec_z) 
\end{split}
\label{eq5}
\end{equation}
where $\theta$ is the normal polar coordinate in the $x-o-y$ plane, and $\phi$ is the angle between vector and the $z$ axis. Assuming the normal is always in the up direction, ${\theta}\in (-\pi, \pi]$ and ${\phi} \in [0, \frac{\pi}{2}]$.
\begin{equation}
  S_{\bm{n}}(ii, jj)= [c_{ii {\Delta\alpha}-\pi}s_{jj{\Delta\alpha}}, s_{ii {\Delta\alpha}-\pi}s_{jj \cdot {\Delta\alpha}},c_{jj {\Delta\alpha}}]
\label{eq6}
\end{equation}
\begin{equation}
  S_{\bm{n}_{g}}(ii', jj')= [c_{ii' {\Delta\alpha}-\pi}s_{jj' {\Delta\alpha}}, s_{ii'  {\Delta\alpha}-\pi}s_{jj' {\Delta\alpha}},c_{jj' {\Delta\alpha}}]
\label{eq7}
\end{equation}
where $ii, ii'=0,1,\cdots,\frac{\pi}{\Delta\alpha}$, and $jj, jj'=0,1,\cdots,\frac{\pi}{2\Delta\alpha}$.

For each $S_{\bm{n}_{cp}}(ii, jj)$, we searched for all $S_{\bm{n}_{g}}(ii', jj')$ satisfying Eq. (\ref{eq3}) in order to create a map $\mathcal{M}:(ii,jj)\to{(ii', jj')}$, which mapped a point normal entry to all $\bm{n}_g$ in the same direction as the normal vector. This map could be generated offline, and this needed to be done only once, thus reducing the computation cost.

Based on $\mathcal{M}$, given the point normals, the feasible candidate $\bm{n}_{g}$ could be rapidly obtained so that gripper orientation samples ($\bm{S}_{O}$) could be generated. Given $\bm{n}_{pcd}$, normals of points located in affordable areas were extracted and azimuthal and polar angles were computed (lines 1-3 in Algorithm 2). The angles were then used to calculate the entry key $\bm{ii}, \bm{jj}$ to query $\mathcal{M}$ to obtain the feasible $\bm{ii'}, \bm{jj'}$, based on which samples ($\bm{S}_{\theta_g}$ and $\bm{S}_{\phi_g}$) of $\theta_g$ and $\phi_g$ were obtained (lines 4-7 in Algorithm 2). Note that only unique $\bm{ii}, \bm{jj}$ values with top-10\% counts were used as entries. This helped to improve the sampling efficiency when the variation in $\bm{n}_{pcd}$ was small. For example, if the input point cloud was set of points in a plane, all $\bm{n}_{pcd}$ and corresponding $\bm{ii}, \bm{jj}$ were the same. Hence, by using unique values, only one unique $\bm{ii}, \bm{jj}$ rather than all $\bm{ii}, \bm{jj}$ were used. As $\bm{n}_g$ depends on only $\theta_g$ and $\phi_g$, $\gamma_g$ could be any value if $\theta_g$ and $\phi_g$ were feasible. Hence, $\gamma_g$ was sampled by the same interval $\Delta\alpha$ in the range $(-\pi, \pi]$ (lines 9-10 in Algorithm 2). The final $\bm{S}_{O}$ were obtained by multiplying the rotation matrix of sampled $\bm{S}_{\theta_g}$, $\bm{S}_{\phi_g}$, and $\bm{S}_{\gamma_g}$.

\begin{algorithm}[tb]
\caption{SampleGripperOrientation}
\label{alg2}
\hspace*{\algorithmicindent} 
$\begin{aligned}
\textbf{Input: }  
&\bm{n}_{pcd}: \textnormal{point normals}\\
&I_{aff}: \textnormal{affordance map} \\
&{\Delta\alpha}: \textnormal{sampling interval}
\end{aligned}$

\hspace*{\algorithmicindent} \textbf{Output: } $\bm{S}_{O}$: gripper orientation samples
\begin{algorithmic}[1]
\State $\bm{n} \gets \bm{n}_{pcd}[I_{aff}>0]$
\State $\bm{\theta} \gets \textrm{arctan}(\bm{n}_y, \bm{n}_x)$
\State $\bm{\phi} \gets \textrm{arccos}(\bm{n}_z) $
\State $\bm{ii} \gets \frac{\bm{\theta}+\pi}{\Delta\alpha}$
\State $\bm{jj} \gets \frac{\bm{\phi}}{\Delta\alpha}$
\State $\bm{ii}, \bm{jj} \gets \textrm{Unique}(\bm{ii}, \bm{jj}) $
\State $\bm{ii'}, \bm{jj'} \gets \mathcal{M}(\bm{ii}, \bm{jj})$
\State $\bm{S}_{\theta_g} \gets \bm{ii'}{\Delta\alpha}-\pi$
\State $\bm{S}_{\phi_g} \gets \bm{jj'}{\Delta\alpha}$
\State $\bm{kk'} \gets 0,1,\cdots,\frac{\pi}{\Delta\alpha}$
\State $\bm{S}_{\gamma_g} \gets \bm{kk'}{\Delta\alpha}-\pi$
\State $\bm{S}_{O} \gets R_z(\bm{S}_{\theta_{g}})R'_y(\bm{S}_{\phi_{g}})R''_z(\bm{S}_{\gamma_{g}})$
\State \Return $\bm{S}_{O}$
\end{algorithmic}
\end{algorithm}

\subsection{Gripper orientation kernel generation and suction cup ID encoding}

\begin{figure}[hbtp]
 \centering
 \includegraphics[keepaspectratio, scale=0.3]{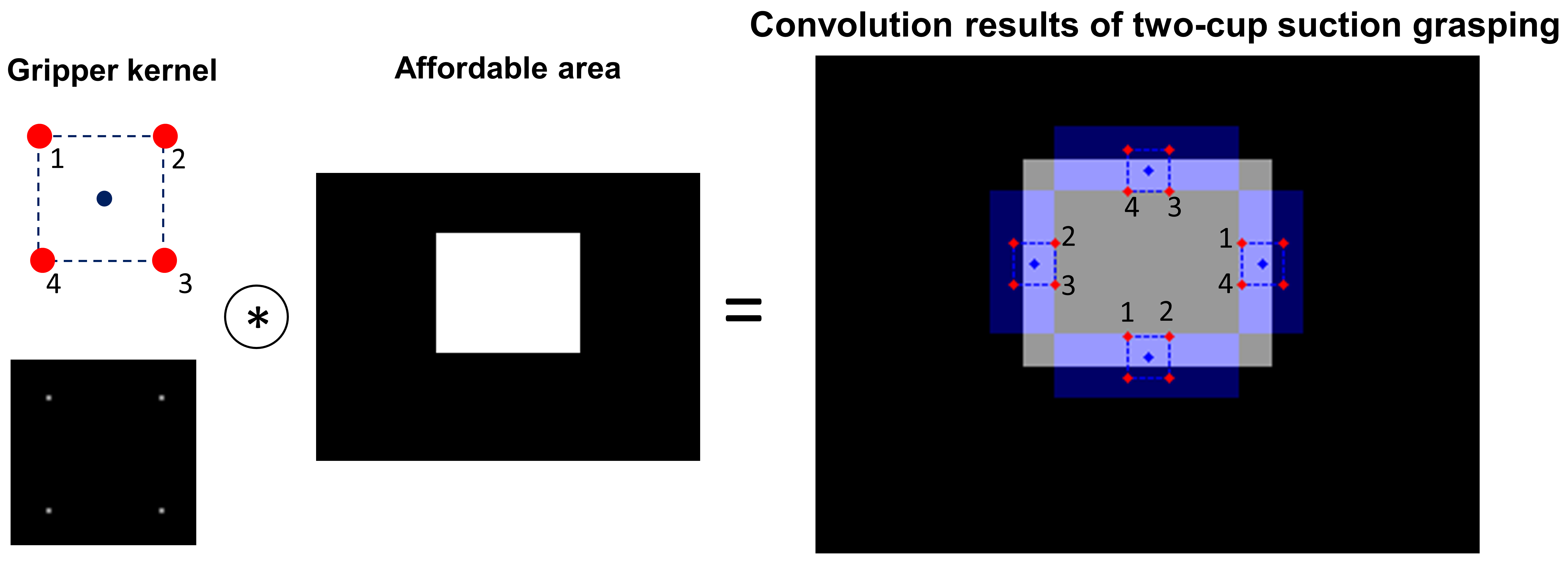}
 \caption{Problem of using a binary kernel. Red dots are cup centers and blue dots are TCP positions. The transparent blue area indicates graspable positions using two cups. The convolution results for two-cup suction grasping are the same for the four cases in which the convolved values for the four cases are all equal to 2. However, although the suction cup center positions in the affordable area are different so that cups to activate differ between the four cases, the activation pattern cannot be directly determined from the convolution result. }
 \label{fig5}
\end{figure}

The kernel representing each candidate gripper orientation generated in Section 5.4 was created for 3D convolution to find the graspable position for each $\bm{S_O}$ as in Algorithm 3. A binary kernel was used to represent gripper poses in previous studies using 2D convolution \cite{40}. However, the convolution results could only determine the graspable position of the kernel, and could not directly determine which suction cup to activate. For example, as shown in Fig. \ref{fig5}, although the convolution results for the four cases are the same, the suction cups to activate are different and cannot be directly determined from the convolution results. Hence we designed a 3D kernel that included the suction cup ID information. Algorithm 3 was used to generate kernels $\bm{S}_{\mathcal{K}}$ of $\bm{S}_O$. The shape of one kernel $\mathcal{K}$ is $N_{\mathcal{K}_x} \times N_{\mathcal{K}_y} \times N_{\mathcal{K}_z}$ where $N_{\mathcal{K}_x}=N_{\mathcal{K}_y}=N_{\mathcal{K}_z}=\lfloor\frac{max(d_{C_i}^0)}{l}\rfloor$. $max(d_{C_i}^0)$ is the max distance of the suction cup center to TCP in gripper local coordinates and $l$ is the grid size of the kernel, which is equal to that of the voxel grid. The kernel indices of cup centers are $\lfloor\frac{C}{l}\rfloor+\lfloor\frac{max(d_{C_i}^0)}{2l}\rfloor$, where $C$ is the cup center positions of $\bm{S}_O$. The kernel grids at cup center kernel indices were filled with encoded vacuum ID information as in line 9 in Algorithm 3. Here, the $i$th suction cup ID information was encoded as $10^{-i}$ such that the cup ID was saved in the $i$th decimal place, and such encoding helped to directly obtain the target suction cups to activate from the decoding convolution results (see Section 5.7).

\begin{algorithm}[tb]
\caption{GeneratedEncodedKernels}
\label{alg3}
\hspace*{\algorithmicindent} 
$\begin{aligned}
\textbf{Input: }  &\bm{S}_{O}: \textnormal{gripper orientation samples} \\
& \bm{C}^0: \textnormal{local cup center positions} \\
&  \textnormal{(see the right side of Fig. \ref{fig2})} \\
& l: \textnormal{voxel size} 
\end{aligned}$

\hspace*{\algorithmicindent} \textbf{Output: } $\bm{S}_{\mathcal{K}}: \textrm{kernels of } \bm{S}_{O}$
\begin{algorithmic}[1]
\State $\textrm{N}_O \gets \textrm{len}(\bm{S}_{O})$
\State $\bm{S}_{\mathcal{K}} \gets \textrm{Zeros}(N_O, N_{\mathcal{K}_x}, N_{\mathcal{K}_y}, N_{\mathcal{K}_z})$
\For{$n \gets 0$ to $\textrm{N}_O$}
\State $\mathcal{K} \gets \textrm{Zeros}(N_{\mathcal{K}_x}, N_{\mathcal{K}_y}, N_{\mathcal{K}_z})$
\For{$i \gets 0$ to $\textrm{N}_c$}
\State $[C[i],1]  \gets  \begin{bmatrix}
\bm{S}_O[n] & \bm{0} \\
\bm{0} & 1
\end{bmatrix}
\begin{bmatrix}
{C^0}[i]^T \\
1
\end{bmatrix}$
\State $C[i] \gets C[i]^T$

\State $\textrm{\# encoding}$
\State $\mathcal{K}[\lfloor\frac{C[i]}{l}\rfloor+\lfloor\frac{max(d_{C_i}^0)}{2l}\rfloor] \gets 10^{-i}$
\EndFor
\State $\bm{S}_{\mathcal{K}}[n] \gets \mathcal{K}$
\EndFor
\State \Return $\bm{S}_{\mathcal{K}}$
\end{algorithmic}
\end{algorithm}

\subsection{3D convolution}

We performed 3D convolution to find the indices in $V$ where the gripper can grasp using multiple suction cups. Because the kernel was generated from an oriented $C^0$ that was located in the same plane as TCP, the corresponding kernel indices of cup centers and TCP were in the same plane, which satisfied Condition 2 in Section 4. Furthermore, as the distances from cup centers to TCP were represented in a kernel scale that was the same as the voxel grid scale, we could slide the kernel over the voxel grid to find the voxel grid index where TCP satisfied Conditions 2 and 4. Specifically, as in Eq. (\ref{eq8}), the kernel was set to each grid cell of the voxel grid to calculate the convolution sum. Note that $N_{\mathcal{K}}$ is the number of kernels, which is equal to $N_O$.

\begin{equation}
 \begin{split}
ConvRes = \sum_{n=0}^{N_{\mathcal{K}}}  \sum_{m=0}^{N_{V_x}}&  \sum_{t=0}^{N_{V_y}}  \sum_{p=0}^{N_{V_z}}  \sum_{i=-\frac{N_{\mathcal{K}_x}}{2}}^{\frac{N_{\mathcal{K}_x}}{2}}  \sum_{j=-\frac{N_{\mathcal{K}_y}}{2}}^{\frac{N_{\mathcal{K}_y}}{2}}  \sum_{k=-\frac{N_{\mathcal{K}_z}}{2}}^{\frac{N_{\mathcal{K}_z}}{2}} \\
&\mathcal{K}[i, j, k] \cdot V[m+i, t+j, p+k]
 \end{split}
\label{eq8}
\end{equation}

\subsection{Convolution results decoding and normal direction check}

Algorithm 4 shows the decode function that decodes the 3D convolution results ($ConvRes$) to generate grasp candidates. As the 3D convolution had the kernel center set to each grid cell of $V$ and then accumulated the kernel values where the corresponding voxel grid value was non-zero (Eq. (\ref{eq8})), the cup to be activated could be determined by obtaining each digit of $ConvRes$. As in line 7 in Algorithm 4, $ConvRes$ was decoded to target $i$th suction cup activation $a_i$ in Eq. (\ref{eq1}) by scaling up $ConvRes$ $10^i$ times and then calculating the value mod 10. If $a_i$ was 1, it indicated that there existed a contact point for the $i$th vaccum cup that should be activated. Otherwise, there was no contact point and the cup should be disabled. For example, for the gripper with two suction cups in Fig. \ref{fig2}, there were four ($2^2$) possible values of convolution results: 0.00, 0.10, 0.01, 0.11, and the decoding result was [0, 0], [1, 0], [0, 1], [1, 1], indicating non-graspable, graspable for only the first cup, graspable for only the second cup, and graspable for both cups, respectively.

As $A$ is a one-hot vector, the sum of $A$ is the number of suction cups to be used. Therefore, we found the indices ($validInd$) of $V$ where the sum of $A$ was greater than or equal to two ($sum(A, dim=-1)\geqslant2$) in order to find the voxel grid indices where multiple suction cups could be used to grasp multiple objects or an object with a large surface. $validInd$ was further converted to TCP positions in world coordinates ($\bm{S}_P$ in Algorithm 4), and the corresponding orientation $\bm{S}_P$, cup center positions $\bm{S}_C$, and target activation status$\bm{S}_A$ could be obtained to generate the grasp candidates ($\bm{G}_{cand}$) as in lines 12-15 in Algorithm 4.

The normal directions of all activated cups ($a_i$=1) of $\bm{G}_{cand}$ were checked to satisfy Condition 3. Specifically, the closest point to the contact point of each activated cup was searched for in $I_{pcd}$, and then the normal of that point was checked for whether it was in the same direction as the gripper axis-z direction by Eq. (\ref{eq3}).

\begin{algorithm}[tb]
\caption{Decode}
\label{alg4}
\hspace*{\algorithmicindent} 
$\begin{aligned}
\textbf{Input: }  &ConvRes: \textnormal{3D convolution results} 
\end{aligned}$

\hspace*{\algorithmicindent} \textbf{Output: } $\bm{G}_{cand}: \textrm{grasp candidates for multiple-cup suction}$
\begin{algorithmic}[1]
\State $\textrm{N}_{\mathcal{K}}, \textrm{N}_{V_x}, \textrm{N}_{V_y}, \textrm{N}_{V_z} \gets convRes.\textrm{shape}$
\State $\bm{A} \gets \textrm{Zeros}(\textrm{N}_{\mathcal{K}}, \textrm{N}_{V_x}, \textrm{N}_{V_y}, \textrm{N}_{V_z}, \textrm{N}_c)$
\For{$Res$ in $ConvRes$}
\For{$i \gets 0$ to $\textrm{N}_c$}
\State $\textrm{\# decoding cup ID}$
\State $\bm{A}[..., i] \gets \lfloor\frac{10^{N_{cup}}}{10^{N_{cup}-i}}Res\rfloor  \bmod 10$
\EndFor
\EndFor
\State $validInd \gets \textrm{sum}(A, \textrm{dim}=-1) \geqslant 2$
\State $\bm{S}_P \gets validInd \cdot l + \bm{B}^{min}$
\State $\bm{S}_O \gets \bm{S}_O[validInd]$
\State $\bm{S}_C \gets \begin{bmatrix}
\bm{S}_O & \bm{S}_P \\
\bm{0} & 1
\end{bmatrix}
\begin{bmatrix}
{C^0}^T \\
1
\end{bmatrix}$
\State $\bm{S}_A \gets \bm{A}[validInd]$
\State $\bm{G}_{cand} \gets [\bm{S}_P, \bm{S}_O, \bm{S}_C, \bm{S}_A]$
\State \Return $\bm{G}_{cand}$
\end{algorithmic}
\end{algorithm}

\subsection{Ranking}

\begin{figure}[hbtp]
 \centering

 \includegraphics[keepaspectratio, scale=0.3]{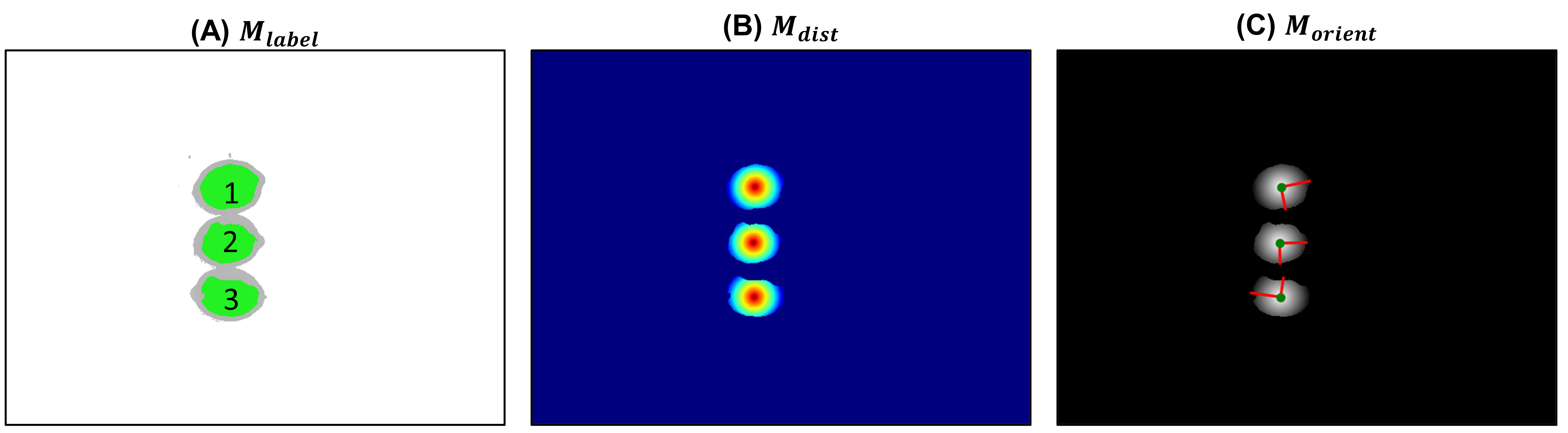}
 \caption{Clustering results. (A) Label map $M_{label}$. (B) Distance map $M_{dist}$. (C) Orientation map $M_{orient}$. }
  \label{fig6}

\end{figure}

Each $\bm{G}_{cand}$ was evaluated and ranked to find the optimal grasp $\bm{G}_{opt}$. We first performed point clustering on the points with non-zero affordance values, which were extracted from $I_{pcd}$ to generate a label map $M_{label}$, distance map $M_{dist}$, and orientation map $M_{orient}$ as shown in Fig. \ref{fig6}.  $M_{label}$ contained the ID label of each cluster and was later used to calculate how many objects could be grasped. $M_{dist}$ contained the 3D distance from each point in the cluster to the cluster center. $M_{orient}$ contained the 3D orientation of each cluster. $M_{dist}$ and $M_{orient}$ were generated for later evaluation of the score ($J$) of $\bm{G}_{opt}$. The height and width of the three maps were the same as those of $I_d$. 

\begin{figure}[hbtp]
 \centering
 \includegraphics[keepaspectratio, scale=0.3]{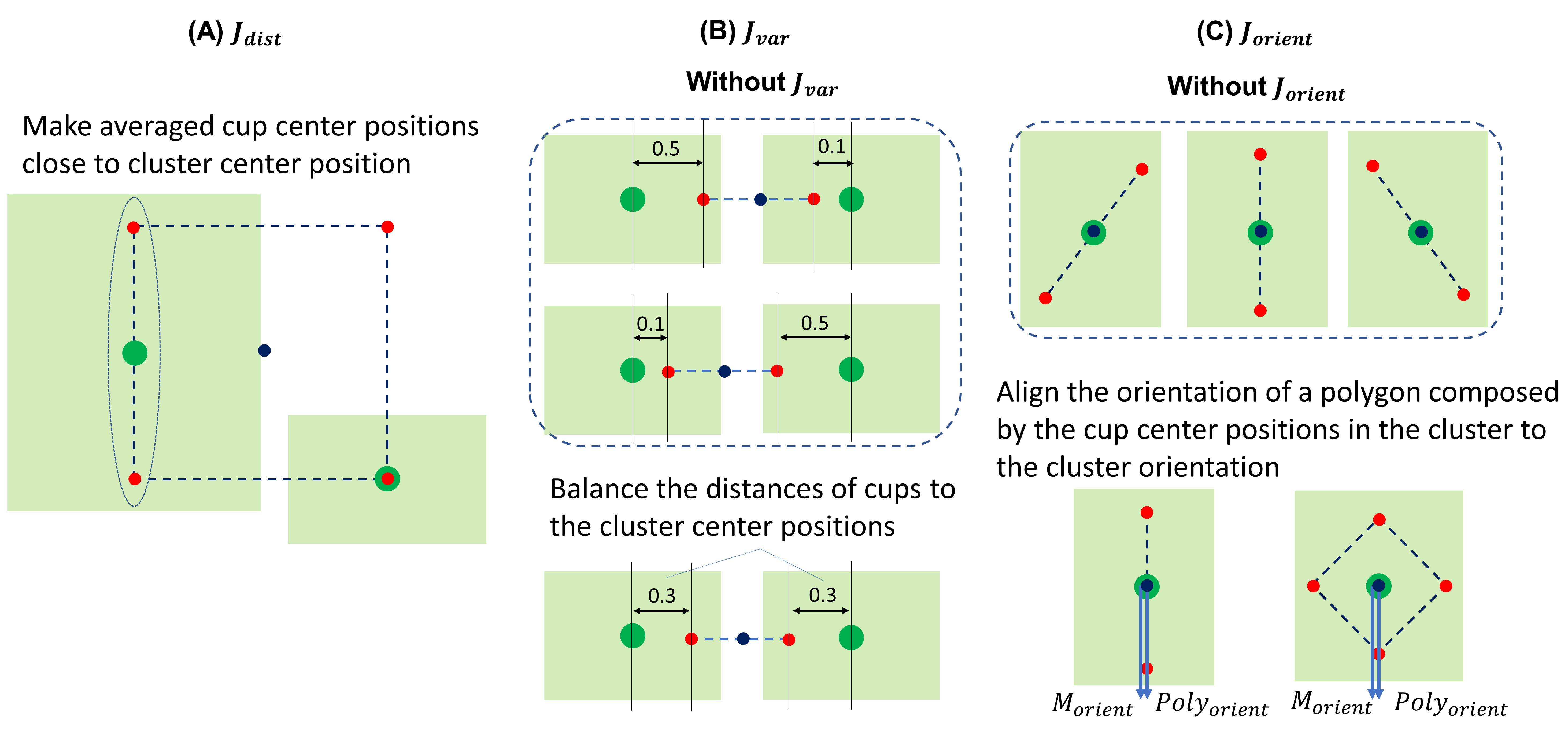}
 \caption{Metrics to evaluate $\bm{G}_{cand}$. Red dots are cup centers and blue dots are TCP positions. Green areas are clusters of affordable areas. Green dots are cluster centers. (A) Distance score $J_{dist}$. (B) Distance variation score $J_{var}$. (C) Orientation score $J_{orient}$. }
 \label{fig7}
\end{figure}

Lines 3-14 in Algorithm 5 evaluated the maximum number of objects that could be grasped $maxObj$ and score $J$ of each $G$ in $\bm{G}_{cand}$ and saved the evaluation results to a dictionary ($rankingRes$). The image coordinates ($\bm{u}_c, \bm{v}_c$) of cup center positions were calculated to obtain the contact point label for each cup in the $M_{label}$. Note that the contact labels might not be unique. If several cups had the same contact point label, it meant that these cups were used to grasp the same object, which had a large surface. If all cups had different contact point labels, it meant that each cup could grasp a unique object. Therefore, the number of unique contact labels was the maximum number of objects that can be grasped by $G$. $J$ was the sum of $J_{center}$, $J_{var}$, and $J_{orient}$. $J_{center}$ evaluated the distance from the cup center or the average of cup centers to the cluster center because it was assumed to be more stable to hold the object at its center. As in Eq. (\ref{eq9}) and Fig. \ref{fig7} (A), $J_{dist}$ was evaluated as the average sum of distances from cups to the cluster center by obtaining the value of $M_{dist}$ at the position of the average cup center position ($\textrm{avg}(\bm{v}_c, \bm{u}_c, contactLabel_i)]$) in each cluster. 

\begin{equation}
J_{dist} = \frac{\sum_{i=0}^{N_{contactLabel}} M_{dist}[\textrm{avg}(\bm{v}_c, \bm{u}_c, contactLabel_i)]}{maxObj}
\label{eq9}
\end{equation}

$J_{var}$ was incorporated because there were cases where one cup was near but another cup was far from the cluster center, and $J_{dist}$ could not evaluate these cases. $J_{var}$ was used to balance the distances of cups to the cluster center positions. Specifically, as in Eq. (\ref{eq10}) and Fig. \ref{fig7} (B), $J_{var}$ evaluated the variance of $M_{dist}$ value at the position of average cup center positions.

\begin{equation}
J_{var} = \frac{\sum_{i=0}^{N_{contactLabel}} (M_{dist}[\textrm{avg}(\bm{v}_c, \bm{u}_c, contactLabel_i)]-J_{dist})^2}{maxObj}
\label{eq10}
\end{equation}

$J_{orient}$ was incorporated to align the orientation of a polygon composed by the cup center positions in the cluster to the cluster orientation. Specifically, we calculated the dot product between the cluster orientation (longer or short axis unit vector) and the polygon orientation as in Eq. (\ref{eq11}) and Fig. \ref{fig7} (C).

\begin{equation}
J_{orient} = \frac{\sum_{i=0}^{N_{contactLabel}} M_{orient}[\bm{v}_c, \bm{u}_c]  \cdot  Poly_{orient}(\bm{v}_c, \bm{u}_c, contactLabel_i)}{maxObj}
\label{eq11}
\end{equation}

The $\bm{G}$ and corresponding $maxObj$ and $J$ were added to the dictionary using $contactLabel$ as a key. Key level (local level) sorting was first performed to sort stored $J$ of $rankingRes[contactLabel]$ (line 15 in Algorithm). Next, dictionary level (global level) sorting was performed to find $\bm{G}_{opt}$ with the highest $maxObj$ and $J$ (lines 16-24 in Algorithm). Note that both $\bm{G}_{opt}$ and sorted $rankingRes$ were returned because if the motion planner failed to find a trajectory to $\bm{G}_{opt}$, it would search for the trajectory to other goals with high $maxObj$ and $J$ in $rankingRes$.

\begin{algorithm}[tb]
\caption{Ranking}
\label{alg5}
\hspace*{\algorithmicindent} 
$\begin{aligned}
\textbf{Input: }  &I_{aff}: \textnormal{affordance map} \\
&I_{pcd}: \textnormal{point cloud} \\
&\bm{G}_{cand}: \textnormal{grasp candidates} 
\end{aligned}$

\hspace*{\algorithmicindent} \textbf{Output: } $\bm{G}_{opt}: \textrm{optimal grasp}$
\begin{algorithmic}[1]
\State $\textrm{M}_{label}, \textrm{M}_{dist}, \textrm{M}_{orient} \gets \textrm{clustering}(I_{aff}, I_{pcd})$
\State $rankingRes \gets \textrm{Dict}()$
\For{$\bm{G}$ in $\bm{G}_{cand}$}
\State $\bm{P}, \bm{O}, \bm{C}, \bm{A} \gets \bm{G}$
\State $\bm{u}_{tcp}, \bm{v}_{tcp} \gets \textrm{getImgCoord}(\bm{P})$
\State $\bm{u}_{c}, \bm{v}_{c} \gets \textrm{getImgCoord}(\bm{C})$
\State $contactLabel \gets \textrm{unique}(M_{label}[\bm{v}_{c}, \bm{u}_{c}])$
\State $maxObj \gets \textrm{len}(contactLabel)$
\State $J_{center} \gets \textrm{calcCenterScore}(M_{dist}, \bm{u}_{c}, \bm{v}_{c}, contactLabel)$
\State $J_{var} \gets \textrm{calcVarScore}(M_{var}, \bm{u}_{c}, \bm{v}_{c}, contactLabel)$
\State $J_{Orient} \gets \textrm{calcOrientScore}(M_{orient}, \bm{u}_{c}, \bm{v}_{c}, contactLabel)$
\State $J \gets J_{center} + J_{var} + J_{Orient}$
\State add $[maxObj, J]$ to $rankingRes[contactLabel]$
\EndFor
\State $rankingRes \gets \textrm{sort}(rankingRes)$
\For{$Res$ in $rankingRes$}
        \If {$maxObj$ in $Res > maxObj$ in $\bm{G}_{opt}$}
            \State $\bm{G}_{opt} \gets Res$
        \ElsIf {$maxObj$ in $Res = maxObj$ in $\bm{G}_{opt}$}
        \If {$J$ in $Res > J$ in $\bm{G}_{opt}$}
            \State $\bm{G}_{opt} \gets Res$
        \EndIf
        \EndIf
\EndFor
\State \Return $\bm{G}_{opt}, rankingRes$
\end{algorithmic}
\end{algorithm}

\section{Experiments}
The multiple-suction-cup planner was validated using previous affordance map datasets as well as real picking experiments. For both validations, thresholds $\varepsilon_1$ in Eq. (\ref{eq3}) and $\varepsilon_2$ in Eq. (\ref{eq4}) were set to $11.5^\circ$ and 0.01 m, respectively. Voxel grid size $l$ was set to 0.005 m. Angle sampling interval$\Delta\alpha$ was set to $5^\circ$. Validations were performed on an Ubuntu 20.04 PC with an 11th Gen Intel Core$^{\textrm{TM}}$ i7-11700K @ 3.60 GHz $\times$ 16 CPU and NVIDIA GeForce RTX 3060 GPU.

\subsection{Validation using a previous affordance map dataset}

We used three datasets to validate the generality of the multiple-suction-cup grasp planner: Suction FCN \cite{9}, SuctionNet-1Billion \cite{38}, and SG-U-Net++ \cite{39}. These datasets provide real RGB-D or synthesized depth images and the corresponding affordance maps. Point clouds converted from depth images and affordance maps in the dataset were used to find the optimal multiple-cup graspable poses and the cups to activate for two-cup and four-cup vacuum grippers. The accuracy of the position and orientation were evaluated by the average error of Eq. (\ref{eq3}) and Eq. (\ref{eq4}), respectively.

\subsection{Validation by picking experiment}

\begin{figure}[hbtp]
 \centering
 \includegraphics[keepaspectratio, scale=0.3]{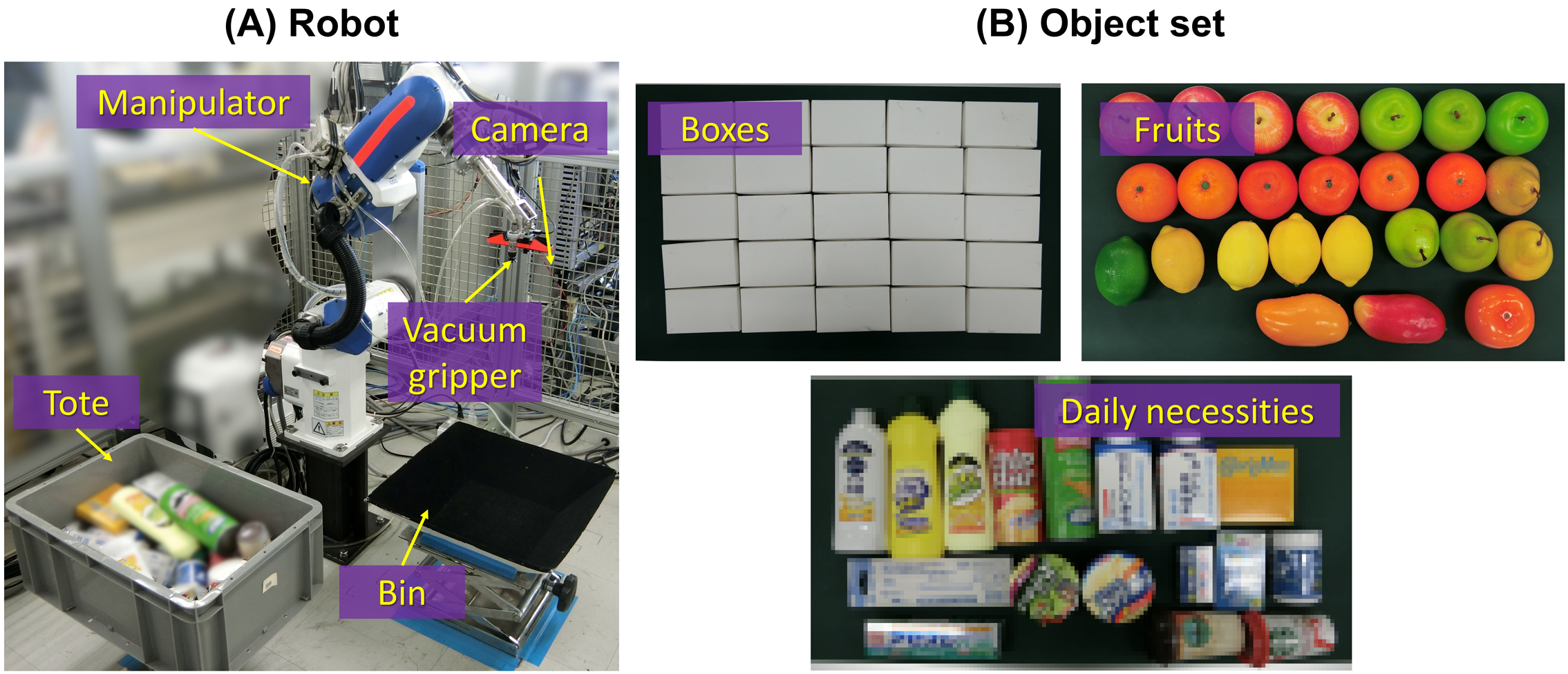}
 \caption{Experiment setup. (A) Robot. (B) Object set.}
 \label{fig8}
\end{figure}

To evaluate the robot picking system and efficiency improvement by using the multiple-suction-cup grasp planner, we conducted picking experiments and compared the results of single-cup (single-object) grasping and multiple-cup (multiple-object) grasping. The robot with a two-cup vacuum gripper was used to pick items from a bin and then place them into a tote (Fig. \ref{fig8} (A)). A camera was installed in the center of the gripper, which captured the depth image at the robot home position. The affordance map was then inferred by SG-U-Net++ based on the depth image. For single-object grasping, the planner in our previous work \cite{39} was used to find the position of maximum affordance value and selected the suction cup that can be used to reach the target grasp point by the shortest trajectory. For multiple-object grasping, a multiple-suction-cup grasp planner was first used to find grasp poses capable of grasping multiple objects or an object with a large surface by using multiple cups. If there was no solution, the planner was switched to the planner for single-object grasping. Trajectories from the home position to grasp poses were generated by MoveIt. As shown in Fig. \ref{fig8} (B), the target object set included boxes, fruits, and daily necessities. The robot was required to pick 50 boxes, 50 fruits, and 51 daily necessities in a cluttered scene. The robot continued grasp attempts until the scene was cleared. A grasp attempt was considered to have failed if the robot could not pick the item or the item was dropped during movement of the manipulator. The results of single-object grasping and multiple-object grasping were evaluated and compared by success rate, picks per hour (PPH), and number of pick attempts. Success rate was defined as the number of successful attempts divided by the number of pick attempts. PPH was defined as the number objects robot could pick in 1 h. The number of pick attempts is defined as number of attempts for the robot to clear the cluttered scene.

\section{Results and discussion}
\begin{figure*}[hbtp]
 \begin{adjustwidth}{-1cm}{}
 \centering
 \includegraphics[keepaspectratio, scale=0.85]{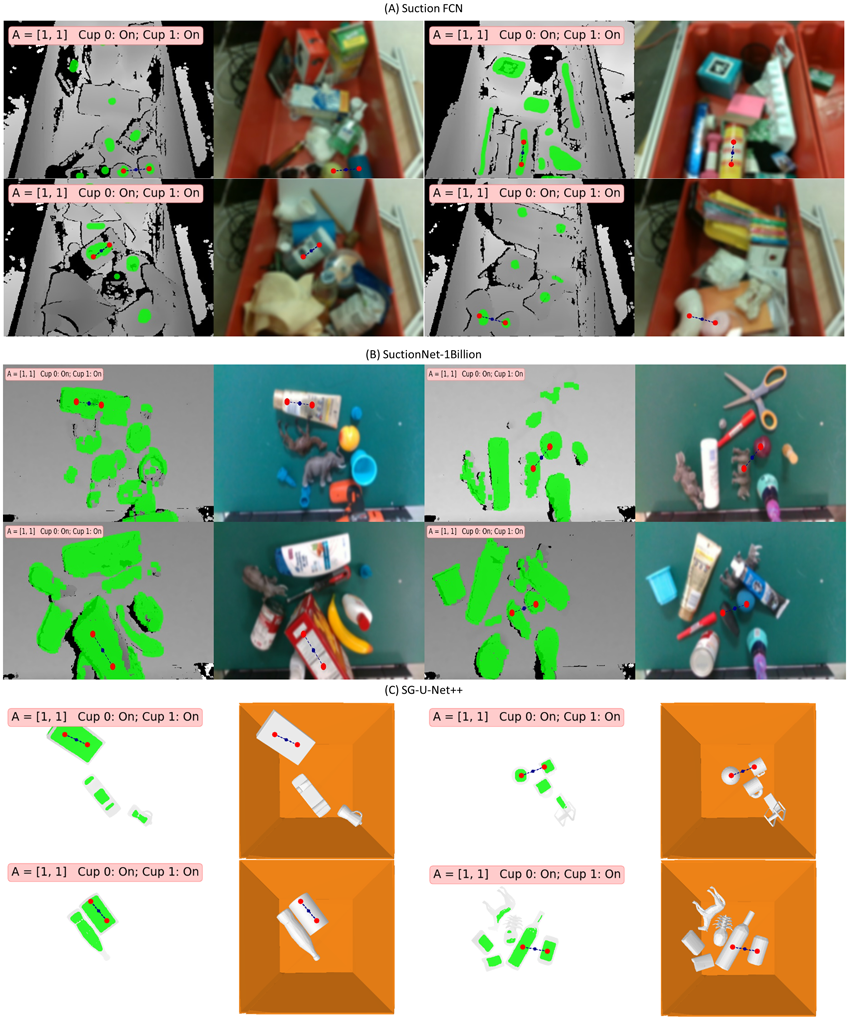}
 \caption{Examples of grasp planning results for two-cup and four-cup vacuum grippers. (A) Suction FCN. (B) SuctionNet-1Billion. (C)SG-U-Net++.}
 
 \label{fig9}
 \end{adjustwidth}
\end{figure*}

\begin{figure*}[hbtp]
 \begin{adjustwidth}{-1cm}{}
 \centering
 \includegraphics[keepaspectratio, scale=0.85]{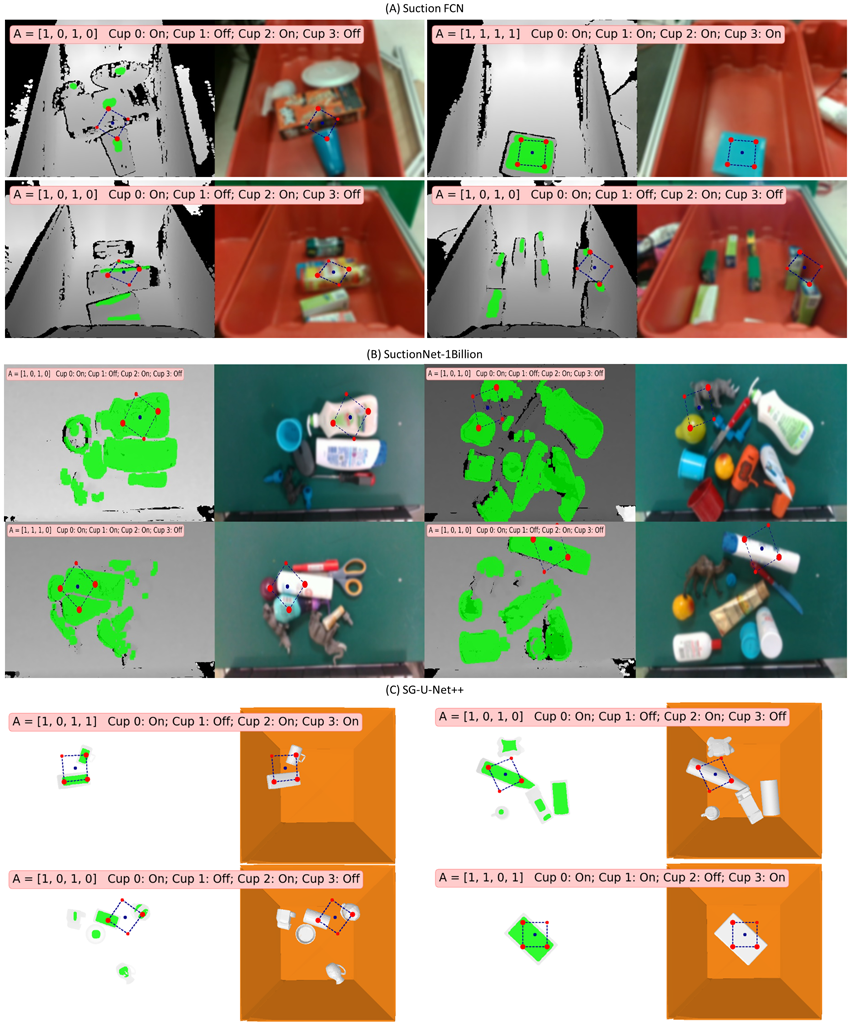}
 \caption{Examples of grasp planning results for two-cup and four-cup vacuum grippers. (A) Suction FCN. (B) SuctionNet-1Billion. (C)SG-U-Net++. Large red spots are centers of activated cups. Small red spots are centers of disabled cups. }
 
 \label{fig10}
 \end{adjustwidth}
\end{figure*}

To our knowledge, this study is the first to propose a grasp planner for multiple-suction-cup grippers to grasp multiple objects or an object with a large surface. Most of the previous studies used a deep neural network to infer the affordance map for finding the optimal grasp for single-cup grasping. Our planner took advantage of the affordance map to find the optimal grasp for multiple-cup grasping. The planner was validated on three previous affordance map datasets and the results are shown in Table \ref{table1}. Our planner successfully found multiple-suction-cup grasps from affordance map from Suction FCN, SuctionNet-1Billion, and SG-U-Net++, indicating the high generality of the planner. There were no significant differences in position orientation error between the two-cup and four-cup gripper planning results. The error was the smallest when grasping was planned based on the affordance map from SG-U-Net++ because SG-U-Net++ used synthesized data (e.g., depth image and point cloud normals) without noise values. Figures \ref{fig9} and \ref{fig10} show examples of grasp planning results for the two-cup and four-cup vacuum grippers. The planner successfully determined which of the cups to activate when grasping.

\begin{table*}
    \centering
    \caption{Position and orientation error of grasp pose}
    \begin{tabular}{ccccc}
        \toprule

        \multirow{3}{*}{Dataset}  & \multicolumn{2}{c}{Two-cup gripper} &\multicolumn{2}{c}{Four-cup gripper}\\
        
& Position Error  & Orientation error & Position error  & Orientation error       \\
& (mean (SD) [m] ) &	(mean (SD) [deg.])& (mean (SD) [m] )&	 (mean (SD) [deg.]) \\
\midrule
Suction FCN     & 6.28$\times10^{-3}$ (0.80$\times10^{-4}$)	&4.50 (6.94)	&5.94$\times10^{-3}$ (0.23$\times10^{-4}$)	&5.04 (11.00)
 \\
suctionNet-1Billion    & 7.67$\times10^{-3}$ (2.43$\times10^{-4}$)	&4.66 (6.30)	&7.64$\times10^{-3}$ (1.41$\times10^{-4}$)	&4.59 (6.98)
\\
SG-U-Net++     & 2.88$\times10^{-3}$ (0.18$\times10^{-4}$)	&2.85 (10.2)	&2.30$\times10^{-3}$ (0.07$\times10^{-4}$)	&2.68 (8.07)
\\
\bottomrule
    \end{tabular}
\label{table1}
\end{table*}

\begin{figure*}[hbtp]
 \centering
 \includegraphics[keepaspectratio, scale=0.85]{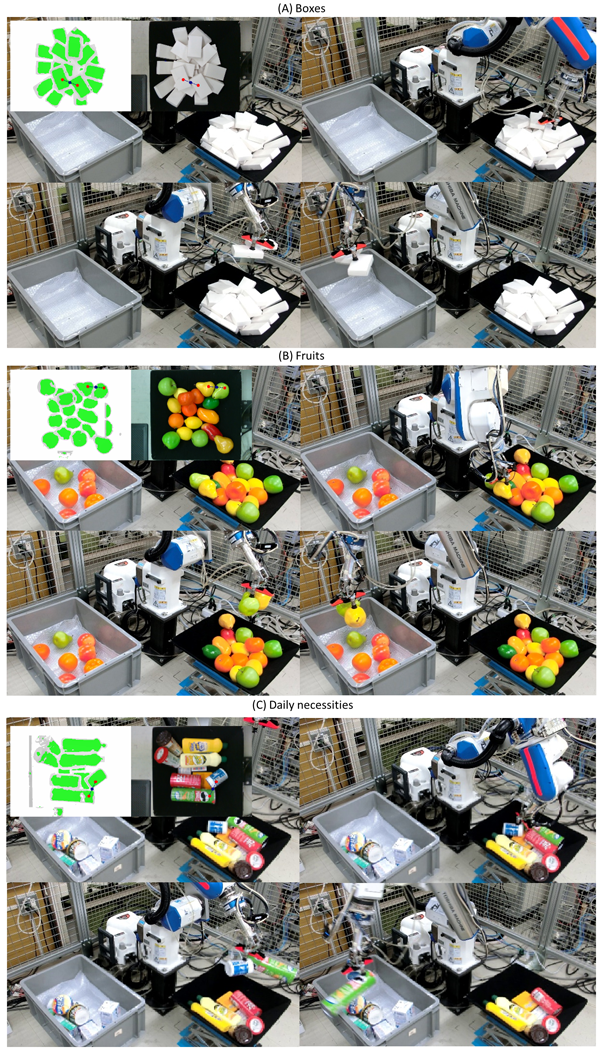}
 \caption{One picking trial for multiple-suction-cup grasp of (A) boxes, (B) fruits, and (C) daily necessities.}
 \label{fig11}
\end{figure*}

The physical experiment results showed that multiple-cup suction grasping can improve the efficiency of picking tasks. Table 2 shows a comparison of experimental results between single-cup (single-object) and multiple-cup (multiple-object) grasping. For single-object grasping, all three object sets could be cleared by the robot. Daily necessities had the highest success rate (91\%) and highest PPH (502) among the three object sets. The success rate of picking fruits was the lowest because the objects had a ball-like shape and rolled and slipped when the gripper pushed them along the normal direction during grasping despite having the correct grasp pose. The success rate of picking boxes was lower than that of daily necessities because when two boxes were very close together, the planner treated them as a single box and grasped the center, which was actually the edge between two boxes. This problem did not occur for the case of multiple-suction-cup grasping because even when two boxes were treated as a single big box, the planner set the averaged cup center positions to the center of the affordable area as shown in Fig. \ref{fig7} so that the cups did not suck the edge between boxes. For multiple-object grasping, all three object sets could also be cleared by the robot. The success rate for grasping boxes (100\%) was the highest among the object sets. The robot picked fruits with the highest speed (PPH=779). Multiple-object grasping improved the picking speed by $1.45\times$ for boxes (PPH: 467 vs. 677), $1.65\times$ for fruits (PPH: 472 vs. 779), and $1.16\times$ for daily necessities (PPH: 502 vs. 583). These results indicated that multiple-suction-cup grasping can improve picking speed. The improvement in picking daily necessities was minor because it was difficult to find multiple-cup graspable poses due to the complicated shapes of the items. Figure \ref{fig11} shows one picking trial for multiple-suction-cup grasping of boxes, fruits, and daily necessities. More trials are shown in the supplementary video file. 

The picking system is expected to be improved in future work aimed at further increasing the picking speed. As described above, one common failure is that objects can move (e.g., roll) after being grasped. We intend to analyze the dynamics (e.g., object shape, friction, and contact force between items) after grasping to find a grasp that moves the object and neighboring objects such that grasp success is improved. Another area for improvement is depth filling because incomplete depth results in low accuracy in estimating the affordance map and normals, and thus leads to low grasp success. Furthermore, we will consider the picking sequence to improve the possibility of picking multiple objects.

\begin{table*}
    \centering
    \caption{Experimental results}
    \begin{tabular}{cccccc}
        \toprule
        Object set              & Method &  Total attempts & Successful attempts & Success rate & PPH  \\
        \midrule
        \multirow{2}{*}{Boxes} & Single-object grasping        & 59        & 50        & 85\%        & 468        \\
                           & Multiple-object grasping        & 36        & 36        &100\%        & 677        \\
        \midrule
        \multirow{2}{*}{Fruits} & Single-object grasping        & 64        & 50        & 78\%        & 472        \\
                           & Multiple-object grasping        & 33        & 31        & 94\%        & 779        \\
        \midrule
        \multirow{2}{*}{Daily necessities} & Single-object grasping        & 56        & 51        & 91\%        & 502        \\
                           & Multiple-object grasping        & 53        & 40        & 75\%        & 583        \\
        \bottomrule
    \end{tabular}
\label{table2}
\end{table*}

\section{Conclusions}
In this study, we proposed a grasp planner for a multiple-suction-cup vacuum gripper. The planner took advantage of an affordance map to find grasp poses for multiple-cup grasping by a 3D convolution-based method. Thanks to the encoded cup ID kernel, the planner could directly determine which cups to activate by decoding the convolution results. The planner exhibited good generality on previous affordance map datasets. The planner also showed the ability to improve picking speed compared with single-cup grasping in physical experiments with a real robot. We will work on improving the planner in future work from several directions including object state analysis after grasping, point cloud or depth image completion, and picking sequence planning.

\bibliographystyle{elsarticle-num} 
\bibliography{references}






\end{document}